\documentclass[9pt,twocolumn]{article}
\advance\topmargin-1in\advance\oddsidemargin-1in
\evensidemargin\oddsidemargin

\makeatletter
\def\@normalsize{\@setsize\normalsize{12pt}\xpt\@xpt
\abovedisplayskip 10pt plus2pt minus5pt\belowdisplayskip \abovedisplayskip
\abovedisplayshortskip \z@ plus3pt\belowdisplayshortskip 6pt plus3pt
minus3pt\let\@listi\@listI}

\def\section{\@startsection {section}{1}{\z@}{20pt plus 2pt minus 2pt}
{8pt plus 2pt minus 2pt}{\centering\normalsize\sc
\edef\@svsec{\thesection.\ }}}
\def\thesection{\Roman{section}}

\def\subsection{\@startsection {subsection}{2}{\z@}{16pt plus 2pt minus 2pt}
{6pt plus 2pt minus 2pt}{\normalsize\sl
\edef\@svsec{\thesubsection.\ }}}
\def\thesubsection{\Alph{subsection}}

\long\def\@makecaption#1#2{
\vskip10pt\begin{center} #1 #2 \end{center}\par\vskip 1pt}
\def\fnum@figure{\raggedright{\footnotesize Fig. \thefigure }.%
\footnotesize}
\def\fnum@table{\footnotesize TABLE \thetable\\\footnotesize\sc}
\def\thetable{\Roman{table}}

\makeatother
\usepackage[bookmarks=false]{hyperref}
\hypersetup{
    colorlinks = true,
    citecolor  = blue,
    linkcolor  = blue,
    urlcolor   = blue,
}
\usepackage[T1]{fontenc}
\usepackage[utf8]{inputenc}
\usepackage{courier}

\usepackage{amsmath}
\usepackage{graphicx}                                      
\usepackage{cleveref}
\usepackage{subfig}
\usepackage{tikz}
\usepackage{pgfplots}
\usepackage{multirow}
\usepackage{booktabs}
\pgfplotsset{compat=newest}
\usepackage{amsfonts}

\newcommand{\m}[1]{\mathbf{#1}}
\graphicspath{{./figs/}}
\Crefformat{figure}{Fig.~#2#1#3}                           
\Crefname{subfigure}{Fig.}{Figs.}
\Crefname{figure}{Fig.}{Figs.}
\Crefformat{table}{TABLE~#2#1#3}                           
\usepackage{titlesec}                                      
\titlespacing*{\section}{0pt}{3.8ex plus .6ex minus .2ex}{0.4ex plus .2ex}
\titlespacing*{\subsection}{0pt}{2.6ex plus .6ex minus .2ex}{0.4ex plus .2ex}

\renewcommand{\vec}[1]{\boldsymbol{#1}}    

\begin{document}
\date{}

\title{\Large\textbf{
VLSI Mask Optimization: From Shallow To Deep Learning
}}

\author{
    Haoyu Yang$^1$,
    Wei Zhong$^2$,
    Yuzhe Ma$^1$,
    Hao Geng$^1$,
    Ran Chen$^1$,
    Wanli Chen$^1$,
    Bei Yu$^1$\\
    $^1$CSE Department, The Chinese University of Hong Kong \\
    $^2$ISE, Dalian University of Technology \\
    {\tt\small \{hyyang,byu\}@cse.cuhk.edu.hk, zhongwei@dlut.edu.cn}
}

\maketitle
\thispagestyle{empty}

{\small\textbf{Abstract---
VLSI mask optimization is one of the most critical stages in manufacturability aware design,
which is costly due to the complicated mask optimization and lithography simulation.
Recent researches have shown prominent advantages of machine learning techniques dealing with complicated and big data problems,
which bring potential of dedicated machine learning solution for DFM problems and facilitate the VLSI design cycle.
In this paper, we focus on a heterogeneous OPC framework that assists mask layout optimization.
Preliminary results show the efficiency and effectiveness of proposed frameworks that have the potential to be alternatives to existing EDA solutions.
}}

\section{Introduction}

VLSI mask optimization is one of the most critical stages in manufacturability aware design, 
which is costly due to the complicated mask optimization and lithography simulation. 
Recent studies have shown prominent advantages of machine learning techniques dealing with complicated and big data problems, 
which bring the potential of dedicated machine learning solution for DFM problems and facilitate the VLSI design cycle \cite{DFM-TCAD2013-Pan,OPC-JJAS2016-Bisschop}.


Related researches include layout hotspot detection \cite{HSD-TCAD2019-Yang,HSD-ASPDAC2019-Yang,HSD-ASPDAC2019-Ye,HSD-ASPDAC2016-Matsunawa,HSD-DAC2013-Yu,HSD-ASPDAC2017-Tomioka,OPC-ASPDAC2019-Jiang}, 
mask optimization \cite{OPC-ASPDAC2019-Geng,OPC-DAC2018-Yang,OPC-ASPDAC2019-Jiang,OPC-SPIE2015-Matsunawa,OPC-TCAD2017-Xu,OPC-DAC2019-Alawieh} and pattern generation \cite{HSD-DAC2019-DeePattern},
all of which contribute to high performance mask optimization flow.
Among the above, 
layout hotspot detection tries to identify regions that are sensitive to process variations and require additional care in OPC stage, 
defect prediction at OPC runtime helps circumvent costly lithography simulation using efficient machine learning engine,
and learning-based mask optimization flows directly speed-up OPC by 
either creating a good mask initialization for legacy OPC engine that requires fewer iterations to converge,
or circumventing costly lithography simulation with regression/classification model and yields faster mask update in each iteration.
These efforts not only bring benefits for modern OPC flow, but also present the importance of legacy OPC engines, which most, if not all, machine learning solutions still rely on.

\begin{figure}
	\centering
	\includegraphics[width=.5\textwidth]{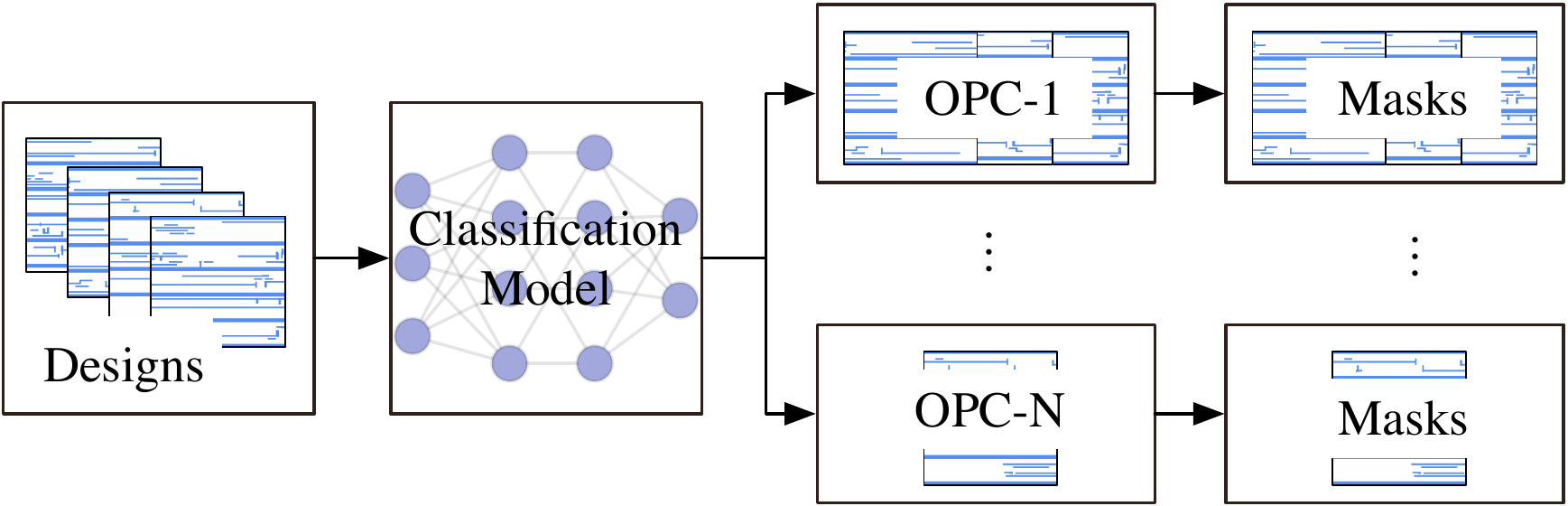}
	\caption{A heterogeneous OPC framework. The classification model identifies whether a design fits different OPC engines.}
	\label{fig:hopc}
\end{figure}

Inverse lithography technique (ILT) \cite{OPC-DAC2014-Gao,OPC-ICCAD2017-Ma,OPC-DAC2018-Yang} and model-based OPC \cite{OPC-TCAD2016-Su,OPC-DATE2015-Kuang} are two representative mask optimization methodologies in literature.
Compared to model-based OPC, ILTs usually promise good mask printability due to larger solution space.
However, the conclusion does not always hold as ILTs require to solve a highly non-convex optimization problem which, sometimes, is hard to converge.
Apparently, different patterns match different OPC engines as can be seen from a simple comparison between \cite{OPC-DATE2015-Kuang} and \cite{OPC-DAC2014-Gao}.
In this paper, we tackle the possibility of machine learning assisting mask optimization from a different perspective,
where a deterministic machine learning model is built to identify a better OPC solution for a given design, as shown in \Cref{fig:hopc}.
This paper makes the following contributions:
\begin{itemize}
	\item We conduct a survey on recent progress of deterministic machine learning models assisting printability estimation and generative models contributing to direct-printable mask synthesis.
	\item We propose a heterogeneous OPC flow where a deterministic machine learning model decides the proper OPC engine for a given pattern.
	\item Experiments show that the proposed framework takes advantage of both ILT and model-based OPC with trivial model prediction overhead.
\end{itemize}

Rest of the paper is organized as follows: 
\Cref{sec:hsd} discusses state-of-the-art researches on layout hotspot detection; 
\Cref{sec:opc} surveys recent progress of OPC and some preliminary machine learning solutions; 
\Cref{sec:hopc} introduce the development of the heterogeneous OPC framework with preliminary experimental results
followed by conclusion in \Cref{sec:conclu}.

\section{Hotspot Detection via Machine Learning}
\label{sec:hsd}
\begin{figure}[tb!] 
	\centering
	\subfloat[]{\includegraphics[width=0.97\linewidth]{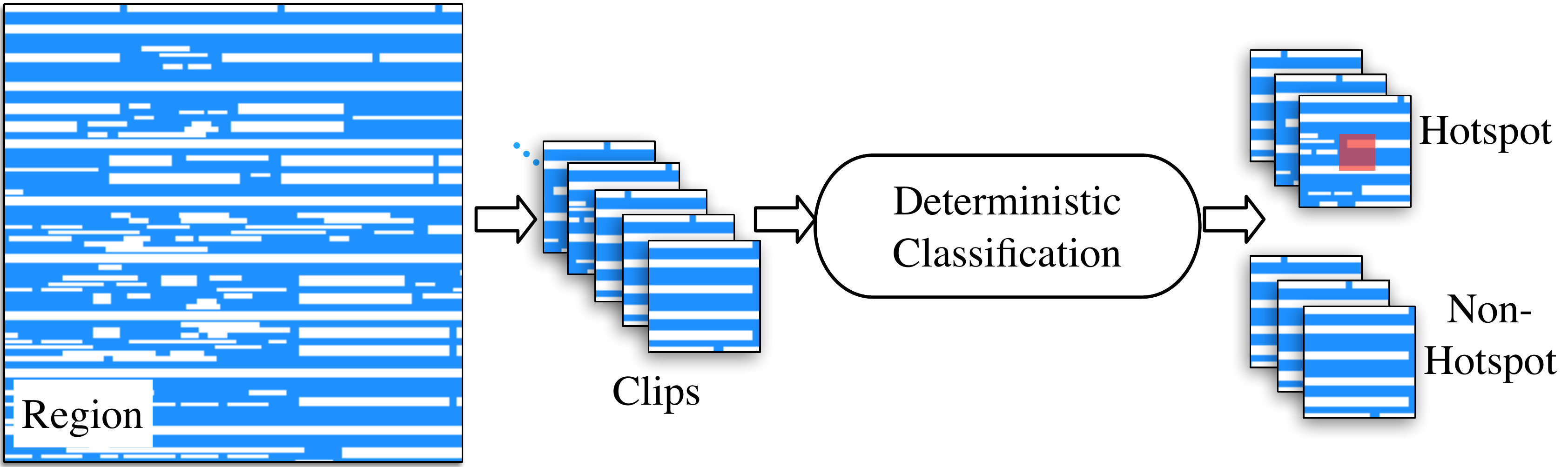} \label
	{fig:flow-a}}\\
	\vspace{-.3in}
	\subfloat[]{\includegraphics[width=0.97\linewidth]{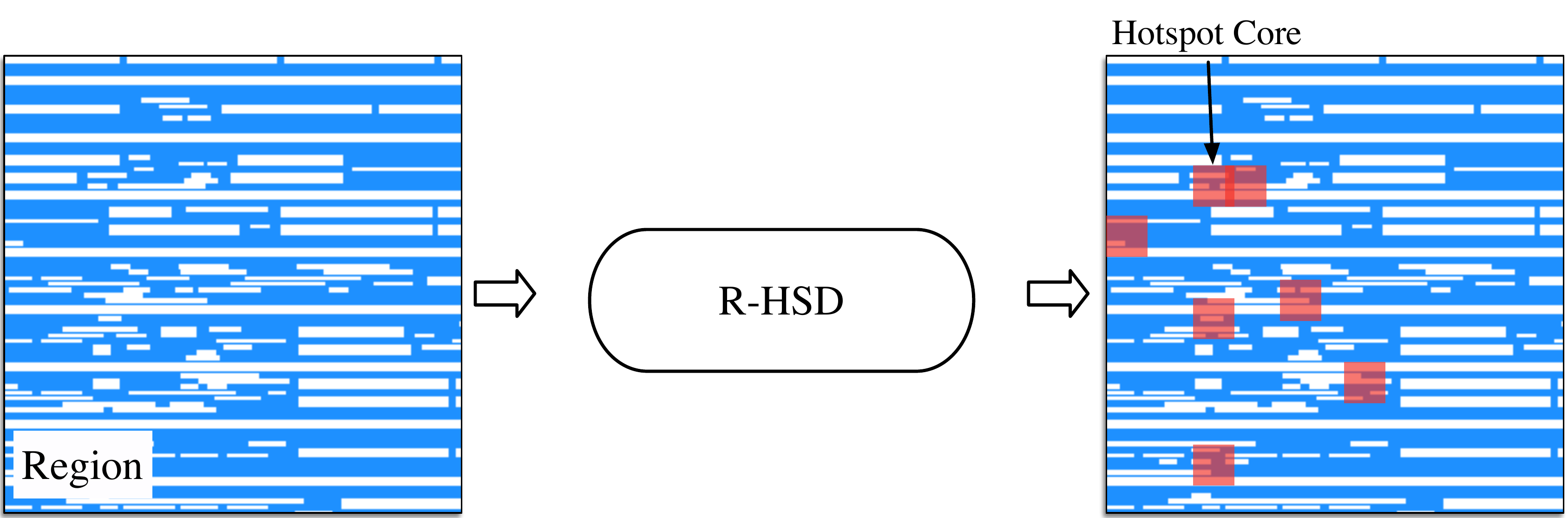} \label
		{fig:flow-b}}
	\caption{(a) Conventional hotspot detection flow vs.~(b) Region based hotspot detection flow as presented in \cite{HSD-DAC2019-Chen}.} 
	\label{fig:flow-compare}
\end{figure}

\subsection{Shallow Machine Learning Solutions}
Before the exploding of deep neural networks, 
traditional machine learning solutions have been deeply investigated to detect lithography hotspots.
Representative solutions include decision tree \cite{HSD-SPIE2015-Matsunawa}, support vector machine (SVM) \cite{HSD-ASPDAC2012-Ding,HSD-TCAD2015-Yu}, artificial neural networks \cite{HSD-ASPDAC2012-Ding} and naive Bayes \cite{HSD-ICCAD2016-Zhang},
which all follows a standard detection flow as in \Cref{fig:flow-a}.

Ding \textit{et al}.~\cite{HSD-ASPDAC2012-Ding} introduce an SVM-based hotspot detection flow, which hierarchically narrows down the search space for hotspot patterns.
Layout designs are converted into to feature space by capturing fragment-based features.
\cite{HSD-TCAD2015-Yu} further enhances the hotspot detection performance using multiple SVM kernels that focus on difference hotspot clusters.
Voting mechanism has made ensemble learning a more promising candidate machine learning framework.
\cite{HSD-SPIE2015-Matsunawa} incorporates Adaboost and decision tree learner for efficient layout hotspot detection and exhibits good trade-off between detection accuracy and false positive penalty.
Another representative ensemble learning framework is proposed in \cite{HSD-ICCAD2016-Zhang},
where the information-theoretic approach is applied in the feature extraction module. 
The problem is solved by a dynamic programming model and embedded into the smooth boosting model with naive Bayes.
The lithography simulation overhead is further reduced.

Different from learning-based model designed for specific manufacturing problem on hotspot detection, 
Jiang \textit{et al}.~\cite{OPC-ASPDAC2019-Jiang} proposed an independent mask printability evaluation framework which detects hotspots caused by EPE.
A second order maximal circular mutual information scheme (SO-MCMI) is presented to select the circle subset.   
The SO-MCMI is formulated as 
\begin{subequations}
	\label{eq:so-mcmi}
	\begin{align}
	\max_{\vec{w}}~~ & \vec{w}^{\top}\vec{M}\vec{w} \\
	\mathrm{s.t.}~~~
	&\sum^{n_c}_{i=1} w_i=n^*_c, w_i \in \{0, 1\}, \forall i,
	\end{align}
\end{subequations}
where $w_i$ in $n_c$-dimensional vector $\vec{w}$ indicates whether the $i^{th}$ circle is selected.
To overcome the potential impacts due to the complicated feature presentations, XGBoost is applied to handle EPE classification and intensity regression modeling.

\subsection{Deep Learning Solutions}

\iftrue
The fast development of deep neural networks brings new opportunities for hotspot detection solutions.
Yang \textit{et al.}~\cite{HSD-TCAD2019-Yang} consider the limitation of conventional machine learning on scalability requirements for printability estimation and feature representation, 
a novel deep learning based hotspot detection model is proposed.
A feature tensor extraction technology is approached to transform origin features into lower scale representations where spatial information is reserved.
To facilitate the training procedure and find a better tradeoffs between accuracy and false alarm, a batch biased learning (BBL) is presented.
BBL adjusts the bias for different instances dynamically which improve the model performance.
The bias function is defined as:
\begin{equation}
    \label{eq:bf}
    \epsilon(l) = \left\{
    \begin{array}{ll}
        \frac{1}{1+\exp (\beta l)}, & \text{if }  l \le 0.3,\\
        0,  & \text{if } l > 0.3,
    \end{array}
    \right.
\end{equation}
where $l$ is the training loss of the current instance or batch in terms of the unbiased ground truth and $\beta$ is a manually determined hyper-parameter that controls how much the bias is affected by the loss.

Adaptive squish pattern is proposed in \cite{HSD-ASPDAC2019-Yang} to handle the multilayer patterns.
Compared with conventional squish patterns presents, 
the adaptive squish pattern not only reserves the property of lossless representation and store layout topologies and geometry information separately in a storage efficient format, 
but also provides a fixed size format which is consistent with most manchine learning models.
To ensure the layout represented by the squish pattern unchanged,
the geometry information $\delta$ should be scaled and duplicated.
To obtain satisfactory $s$ to change the topology matrix to a desired
size as well as attaining low variance $\delta$,
the problem can be formulated as 
\begin{subequations}
    \label{eq:prob1}
    \begin{align}
        \min_{\vec{s}}~~ & ||\vec{\delta}^\prime||_{\infty} \\
        \text{s.t.~~~}
        &\delta_i^\prime = \delta_i / s_i, \forall i, \\
        &s_i \in \mathbb{Z}^{+}, \forall i, \\
        &\sum_{i} s_i=d,
    \end{align}
\end{subequations}
where the geometry information before and after scaling are denoted as $\vec{\delta}$ and $\vec{\delta}^\prime$.
Gradient vanishing problem during the training is also considered and a specific residual convolution block is used to enhance the performance.

Imbalance of positve and negative samples of layout patterns are crital problem especially in machine learning based methods. 
A robust performance metric is needed to evaluate the model performance.
ROC curve based measure for hotspot detection algorithm is proposed in \cite{HSD-ASPDAC2019-Ye}, 
which provides a holistic view of imbalance on hotspot detection dataset.
Multiple loss functions for neural network models are applied to handle the imbalance problem during training.
A general loss function designed for maximize the AUC score can be expressed as
\begin{equation}
    \mathcal{L}_{\Phi}(f)=\frac{1}{N_{+} N_{-}} \sum_{i=1}^{N_{+}} \sum_{j=1}^{N_{-}} \Phi\left(f\left(\vec{x}_{i}^{+}\right)-f\left(\vec{x}_{j}^{-}\right)\right),
\end{equation}
where $f(\vec{x}_{i}^{+})$ and $f(\vec{x}_{j}^{-})$ are the prediction output of positive and negative samples of model $f$ respectively.
$N_{+}$ and $N_{-}$ are number of positive and negative samples. 
The new loss functions present in \cite{HSD-ASPDAC2019-Ye} outperform the traditional cross-entropy
loss on the state-of-the-art neural network model.

While these works deal with the patterns in small clips, 
the large regions with multiple hotspots cannot be handled directly.
Recently, a region based method proposed by Chen \textit{et al.}~\cite{HSD-DAC2019-Chen} solve this problem by enlarging the small clip into large regions (as depicted in \Cref{fig:flow-b}).
Inspired by the object detection task in computer vision field, 
a regression and classification multi-task framework is designed to handle multiple hotspots in large regions in a single epoch.
The clip proposal network is applied to sample hotspot and non-hotspot regions for both classification and regression training.
The loss function for regression on clip $i$ can be written as 
\begin{align}
    l_{loc}(l_i , l'_i) = \left \{
        \begin{aligned}
            &\frac{1}{2}(l_i-l'_i)^2, \ \ && \textrm{if } |l_i - l'_i|<1, \\  
            &|l_i - l'_i|-0.5,              && \textrm{otherwise}, 
        \end{aligned}
        \right .
\end{align}
where $l_{i}$ and $l'_{i}$ are the coordinates of prediction and ground truth respectively. 
The classification loss for clip $i$ can be formulated as
\begin{align}
    l_{hotspot}(h_i,h_i^{'})=-(h_i \log h_{i}^{'} + h_{i}^{'} \log h_i),
\end{align}
where $h_i$ is the prediction of the model and $h'_i$ is the label.
Compared to the deterministic classification flow,
the performance in \cite{HSD-DAC2019-Chen} got improved greatly.

\subsection{Overcome Imbalance: Pattern Generation}
\label{sec:dptn}

In real VLSI manufacturing scenario, hotspot patterns are usually fetal but rare in a design.
This brings challenge for most learning-based solutions which require massive and diverse hotspot data to get a machine learning model well trained.
\cite{HSD-DAC2019-DeePattern} studies the possibility of generating DRC-clean test layout patterns with a generative machine learning model called transforming convolutional auto-encoder (TCAE).
Derived from transforming auto-encoder (TAE) \cite{AE-ICANN2011-Hinton}, TCAE replaces capsule units with simpler latent vector nodes to represent part-whole feature representation.
The identity mapping in TCAE-training allows a neural network to capture certain design rules.
Dedicated perturbations on latent vectors create diverse and DRC-clean patterns.

\fi

\section{Mask Optimization via Machine Learning}
\label{sec:opc}

\begin{figure}[h!]
	\subfloat[]{\includegraphics[width=1.0\linewidth]{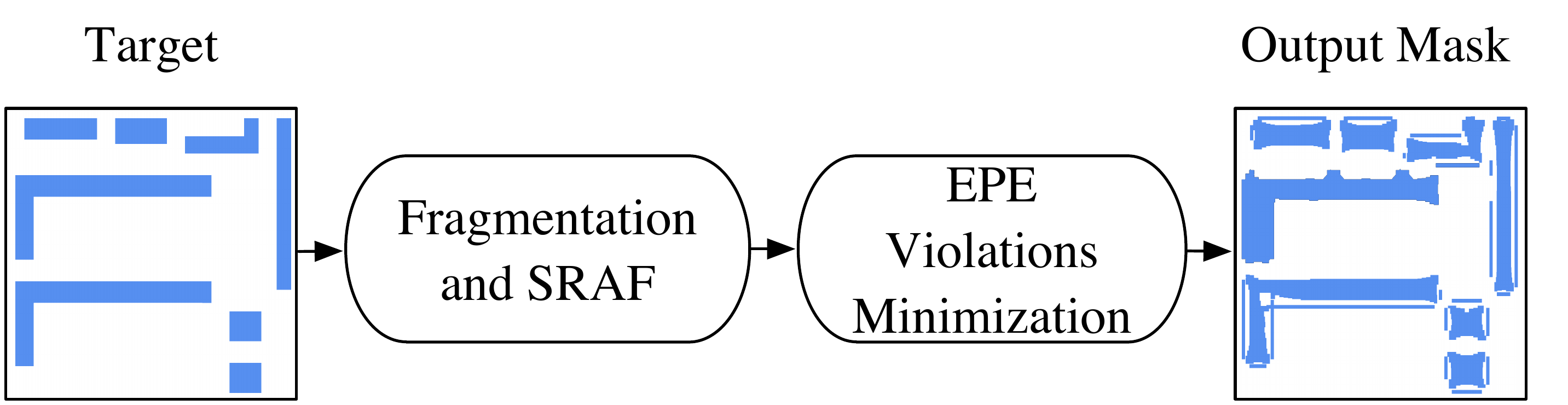}} \label{fig:opc-model}\\
    \vspace{-.1in}
	\subfloat[]{\includegraphics[width=1.0\linewidth]{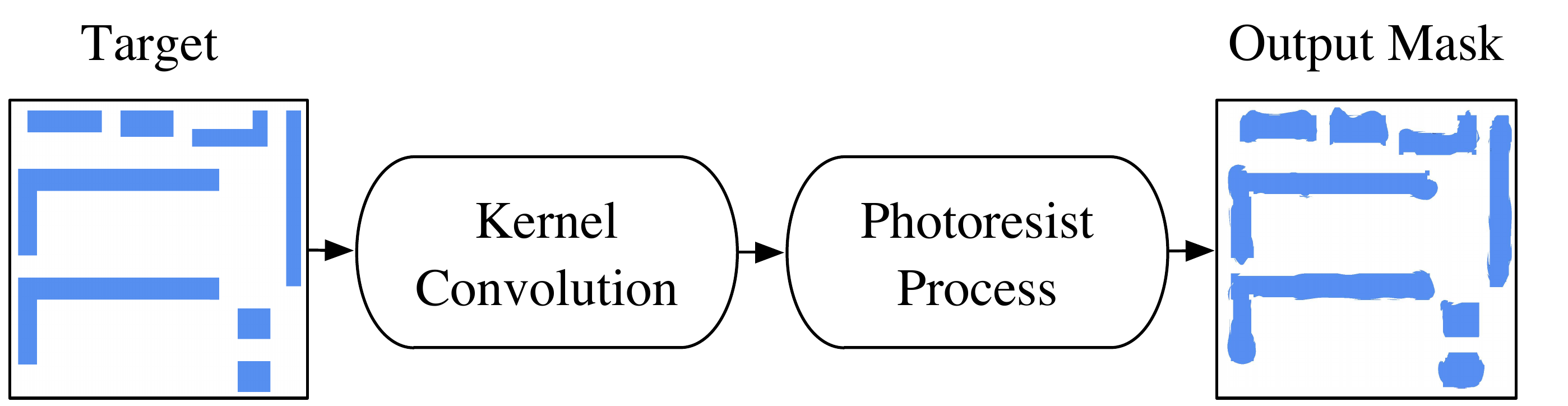}} \label{fig:opc-ilt}
	\caption{The flows of conventional OPC approaches: (a) model-based; (b) ILT-based.}
	\label{fig:opc-all}
\end{figure}

Mask optimization ensures good mask printability and hence improves chip manufacturing yield. 
In advanced technology nodes, the conventional mask optimization processes including model-based and ILT-based approaches consume increasingly more computational resources. 
The flows of model-based and ILT-based approaches are shown in \Cref{fig:opc-all}.
In this section, we will discuss several machine learning-based alternatives that assist traditional mask optimization flow.

\subsection{Machine Learning-based OPC}
The superiority of machine learning-based solutions has been evaluated in OPC \cite{OPC-SPIE2015-Matsunawa}. 
However, the lack of scalability under advanced technology nodes becomes the main issue hindering the widespread deployment of a model-based OPC framework.
Aiming at addressing the scalability issue, a fast machine learning-based mask printability prediction (MPP) framework \cite{OPC-ASPDAC2019-Jiang} for lithography-related applications has bee proposed. 
What's more, the work can be extended to improve the scalability for different lithography-related applications. 
To enable the performance of the machine learning-based flow, a matrix-based concentric circle sampling (MCCS) method and a second-order circle subset selection algorithm for feature extraction are designed in \cite{OPC-ASPDAC2019-Jiang}.
The MPP framework has been demonstrated its effectiveness by being applied to a conventional mask optimization tool. 

Existing machine learning models \cite{OPC-JOIOP2013-Luo,OPC-TSM2008-Gu,OPC-SPIE2015-Matsunawa} can only perform pixel-wise or segment-wise mask calibration that is not computationally efficient. 
In accordance with the critical problem, \cite{OPC-DAC2018-Yang} proposes a generative adversarial network (GAN) based mask optimization flow that takes target circuit patterns as input and generates quasi-optimal masks for further inverse lithography technique (ILT) refinement. 

To enhance the computational efficiency and alleviate the over-fitting issue, training topologies are synthesized. 
For a faster training procedure, an ILT-guided pre-training flow is proposed in \cite{OPC-DAC2018-Yang} to initialize the generator with intermediate ILT results. 
Besides, the authors design new objectives of the discriminator to ensure the model is trained toward a target-mask mapping instead of a distribution. 
The new objective function is as follows:
\begin{align*}
    \min_{\vec{G}} & \max_{\vec{D}} \ \mathbb{E}_{\vec{Z}_t\sim \mathcal{Z}}[1-\log (\vec{D}(\vec{Z}_t, \vec{G}(\vec{Z}_t))) \\
    &+ ||\vec{M}^\ast- \vec{G}(\vec{Z}_t)||_n^n] + \mathbb{E}_{\vec{Z}_t\sim \mathcal{Z}}[\log (\vec{D}(\vec{Z}_t, \vec{M}^\ast))],
\end{align*}
where $\vec{Z}_t$ represents the target layout, 
$\m{G}$ for the generator output, 
$\vec{D}$ for the discriminator output, $p_x$ for some distribution, 
$\vec{M}^\ast$ for the reference mask, 
and a set of target patterns $\mathcal{Z}=\{{\vec{Z}_{t,i}}, i=1,2,\dots,N\}$ and a corresponding reference mask set $\mathcal{M}=\{\vec{M}^\ast_i, i=1,2,\dots,N\}$.
Experimental results have verified that this flow can facilitate the mask optimization process as well as ensure a better printability.

\subsection{Machine Learning-based SRAF Insertion}
Although conventional OPC can size the mask to give the correct critical dimension (CD) on the wafer, it cannot make the isolated target pattern become dense \cite{OPC-SST2003-Mack}. 
As a result, sub-resolution assist feature (SRAF) \cite{OPC-USP2009-Wallace} insertion is proposed. 
There is a wealth of literature on the topic of SRAF insertion for mask optimization, which can be roughly divided into three categories: 
rule-based approach, model-based approach, and machine learning-based approach. 
However, prior machine learning-based approaches \cite{OPC-ISPD2016-Xu,OPC-TCAD2017-Xu} lack well-discrimination feature extraction techniques as well as a global view in SRAF designs, which leads to unsatisfied simulation results. 

Geng \textit{et al}.~firstly revise conventional concentric circle area sampling (CCAS) feature construction method, by proposing a supervised online dictionary learning algorithm for simultaneous feature extraction and dimensionality reduction \cite{OPC-ASPDAC2019-Geng}. 
In other words, label information is not only utilized in learning stage but also imposed in feature extraction stage, which in turn benefits the learning counterpart. 
\Cref{eq:sraf} is the main objective function for supervised feature revision,
where $ \vec y_t \in \mathbb {R}^{n} $ refers to an input CCAS feature vector, 
$\vec q_t \in {\mathbb {R}^s}$ for discriminative sparse code of $t$-th input feature vector, 
$h_t \in {\mathbb {R}}$ for the label of input, 
$\vec x_t \in \mathbb{R}^{s}$ for sparse codes, 
$ \vec D = \left\{ {{\vec d_j}} \right\}_{j = 1}^s, {\vec d_j} \in {\mathbb{R}^{n}}$ for the dictionary made up of atoms to encode input features, 
$\vec A \in {\mathbb {R}^{s \times s}}$ for a matrix transforming original sparse code $\vec x_t$ into discriminative sparse code, 
$\vec W \in {\mathbb {R}^{1 \times s}}$ the related weight vector, 
$\alpha$ and $\beta$ for the balancing hyper-parameters.
\begin{align}
    \label{eq:sraf}
    \small
    \min_{ \vec{x}, \vec{D}, \vec{A}, \vec{W}} \dfrac{1}{N}\sum\limits_{t= 1}^N  & \{ \dfrac{1}{2}
    \left\| \left ({\vec y^\top_t}, \sqrt{\alpha} {\vec q^\top_t}, \sqrt{\beta}{h_t} \right ) ^{\top} -
    \begin{pmatrix}
        \vec D               \\
        \sqrt{\alpha} \vec A \\   
        \sqrt{\beta} \vec W
    \end{pmatrix}
    {\vec x_t} \right\|_2^2 \nonumber \\
    & + \lambda {{\left\| {{\vec x_t}} \right\|}_p} \}.
\end{align}

To consider SRAF design rules in a global view, the authors construct an integer linear programming (ILP) model in the post-processing stage of their SRAF insertion framework. 
Experimental results demonstrate the efficacy of the proposed SRAF insertion flow in \cite{OPC-ASPDAC2019-Geng}.
 
\begin{figure}
    \vspace{-.2in}
    \centering
    \subfloat[]{\includegraphics[width=.38\linewidth]{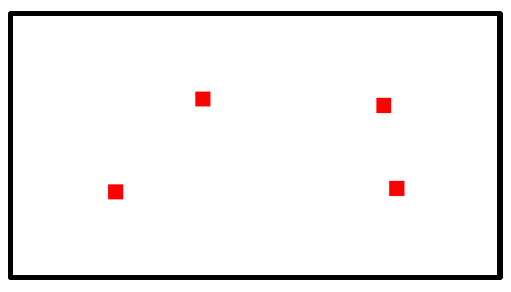} \label{fig:SRAF-a}}
    \subfloat[]{\includegraphics[width=.38\linewidth]{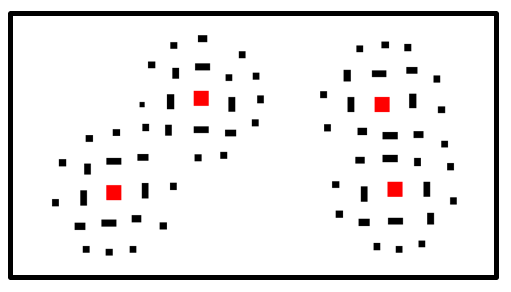} \label{fig:SRAF-b}} \\
    \includegraphics[width=.418\linewidth]{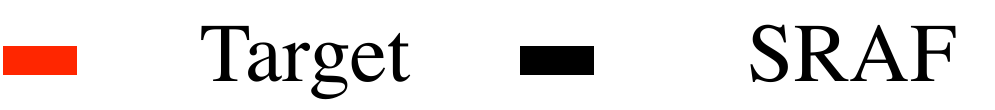}
    \caption{The visualization of the SRAF image translation in \cite{OPC-DAC2019-Alawieh}: (a) Original layout with target contacts; (b) SRAFed layout.}
    \label{fig:sraf}
\end{figure}

However, \cite{OPC-ASPDAC2019-Geng} lies on raw CCAS feature which is manually-crafted but not automatically learnt by the learning model yet. 
Besides, the grid-based ILP method lacks efficiency, especially for large designs.
So there still exists big room to improve. 
Very recently, GAN-SRAF \cite{OPC-DAC2019-Alawieh} casts the original SRAF insertion as an image-to-image translation problem where a layout is translated from its original domain to SRAFed layout domain. 
The visualization of the SRAF image translation is shown in \Cref{fig:sraf}.
To achieve this formulation, Alawieh \textit{et al}.~firstly adopt conditional generative adversarial network (CGAN) in SRAF insertion. 
In addition, to fit CGAN training, a novel multi-channel heatmap encoding/decoding scheme is proposed to map layouts to images without information loss. 
The loss function is designed as \Cref{eq:gansraf}: 
\begin{align}
   \label{eq:gansraf}
    \min _{G} \max _{D} \ & \mathbb{E}_{\vec x, \vec y}[\log D(\vec x, \vec y)]  + \mathbb{E}_{ \vec x,  \vec z}[\log (1-D(\vec x, G(\vec x, \vec z)))] \nonumber \\
						  & +\lambda_{L 1} \mathbb{E}_{\vec x, \vec z, \vec y}\left[\|\vec y-G(\vec x, \vec z)\|_{1}\right], 
\end{align}
where $\vec x$ is an observed image,  $\vec y$ an output image, $\vec z$ a random noise vector. 
$G$ and $D$ refer to the generator and discriminator in CGAN respectively. 
To further reduce blurring, the authors adopt $L1$-norm rather than $L2$-nom. 
With comparable lithographic performance, GAN-SRAF framework surpasses prior works on insertion speed.

\subsection{OPC in Multiple Patterning Scenarios}

In advanced technology nodes, layout decomposition and mask optimization are two of the most critical RET stages. 
In layout decomposition, a target image is divided into several masks, 
while in mask optimization, each decomposed mask is optimized by some RET techniques like OPC \cite{TPL-TCAD2015-Yu}.

\cite{OPC-JM3-2007-Poonawala} is a pioneer work that considers multiple exposure effects in ILT framework.
To automatically synthesize the masks and then print the desired wafer pattern, \cite{OPC-JM3-2007-Poonawala} first combines ILTs and double-exposure lithography.
Via inverting the forward model from mask to wafer, ILTs synthesize the input mask to obtain the required wafer pattern. 
On the other hand, double-exposure lithography exploits two masks under two illumination settings to print the desired wafer pattern. 
The objective function of \cite{OPC-JM3-2007-Poonawala} is shown in \Cref{eq:mp1}, which is formulated as minimizing the $L2$-norm of the difference between the desired pattern $\vec{z}^{*}$ and the aerial image $|\vec{H} \vec{a}|^{2}+|\vec{H} \vec{b}|^{2}$. $\vec{H}$ is a jinc function with cutoff frequency $NA/\lambda$, and $\vec{a}$, $\vec{b}$ are sampled from two input masks.
\begin{equation}
	\min _{\vec{a}, \vec{b}} F(\vec{a}, \vec{b})=\underset{\vec{a}, \vec{b}}{\operatorname{argmin}}\left\|\vec{z}^{*}-|\vec{H} \vec{a}|^{2}-|\vec{H} \vec{b}|^{2}\right\|_{2}^{2}. 
	\label{eq:mp1}
\end{equation}

However, \cite{OPC-JM3-2007-Poonawala} has not addressed the layout decomposition problem yet. 
Ma \textit{et al}.~firstly develops a unified optimization framework which solves layout decomposition and mask optimization simultaneously \cite{OPC-ICCAD2017-Ma}.
To compatible with the objective,  an unified mathematical formulation $\min_{\vec{M}_{1}, \vec{M}_{2}} F =\left\|\vec{Z}_{t}-\vec{Z}\right\|_{2}^{2}$ is proposed in \cite{OPC-ICCAD2017-Ma},
where $\vec{Z}_{t}$ represents the target image with $\vec{Z}$ the printed image, $\vec{M}_{1}$ and $\vec{M}_{2}$ for output masks.
A gradient-based optimization approach with a set of discrete optimization techniques is also proposed to solve the problem efficiently. 
The experimental results in \cite{OPC-ICCAD2017-Ma} demonstrate the efficacy of the unified framework.

\section{Heterogeneous OPC}
\label{sec:hopc}
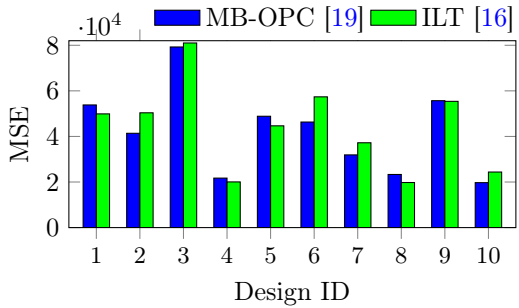
\begin{figure}[tb!]
	\centering
	\begin{tikzpicture}
\pgfplotsset{
	width =0.418\textwidth,
	height=0.226\textwidth,
	every axis plot/.append style = {font = \small},
	/pgfplots/bar cycle list/.style={/pgfplots/cycle list={%
			{black,fill=blue},
			{black,fill=green},
		}
	},
}
\begin{axis}[
ybar=0pt,
enlarge x limits=0.07,
bar width=5pt,
legend style={at={(0.6,1.0)},
	draw=none,anchor=south,legend columns=2},
area legend,
ylabel={MSE},
ymin = 0,
ymax = 82000,
symbolic x coords={1,2,3,4,5,6,7,8,9,10},
xtick=data,
ytick={0,20000,40000,60000,80000},
ylabel near ticks,
x tick label style={font=\small},
xlabel={Design ID}
]
\addplot  coordinates {(1,53816) (2,41382) (3,79255) (4,21717) (5,48858)(6,46320)(7,31898)(8,23312)(9,55684)(10,19722)};
\addplot  coordinates {(1,49893) (2,50369) (3,81007) (4,20044) (5,44656)(6,57375)(7,37221)(8,19782)(9,55399)(10,24381)};
\legend{MB-OPC \cite{OPC-DATE2015-Kuang},ILT \cite{OPC-DAC2014-Gao}}
\end{axis}

\end{tikzpicture}
	\caption{Performance gap between model-based OPC and ILT on ten designs from ICCAD2013 CAD Contest \cite{OPC-ICCAD2013-Banerjee}.}
	\label{fig:mse-gap}
\end{figure}


\begin{table}[tb!]
	\centering
	\caption{Evaluation of the proposed H-OPC.}
	\label{tab:rst}
	\renewcommand{\arraystretch}{1.0}
    \resizebox{8.8cm}{!} {
        \begin{tabular}{c|cc|cc|cc}
            \toprule
            \multirow{2}{*}{ID} & \multicolumn{2}{c|}{MB-OPC \cite{OPC-DATE2015-Kuang}} & \multicolumn{2}{c|}{ILT \cite{OPC-DAC2014-Gao}} & \multicolumn{2}{c}{H-OPC} \\
            & MSE & Time     & MSE  & Time     & MSE & Time    \\ \midrule
            1         & 53816        & 278          & 49893      & 1280        & 49893       & 1280        \\
            2         & 41382        & 142          & 50369      & 381         & 41382       & 142         \\
            3         & 79255        & 152          & 81007      & 1123        & 79255       & 152         \\
            4         & 21717        & 307          & 20044      & 1271        & \underline{21717}       & \underline{307}       \\
            5         & 48858        & 189          & 44656      & 1120        & 44656       & 1120        \\
            6         & 46320        & 353          & 57375      & 391         & 46320       & 353         \\
            7         & 31898        & 219          & 37221      & 406         & 31898       & 219         \\
            8         & 23312        & 99           & 19782      & 388         & 19782       & 388         \\
            9         & 55684        & 119          & 55399      & 1138        & \underline{55684}       & \underline{119}      \\
            10        & 19722        & 61           & 24381      & 387         & 19722       & 61          \\ \midrule
            Avg.      & 42196.4      & \textbf{191.9}        & 44012.7    & 788.5       & \textbf{41030.9}     & 414.1       \\ 
            Ratio     & 1.03         & \textbf{0.46}
                      & 1.07         & 1.90
                      & \textbf{1.00}& 1.00  \\ \bottomrule
        \end{tabular}
    }
\end{table}

Previous works have shown that different OPC engines exhibit advantages on different designs.
\cite{OPC-DAC2014-Gao} and \cite{OPC-DATE2015-Kuang} are two representative implementations of ILT and model-based OPC engine.
\Cref{fig:mse-gap} depicts the performance gap of two engines on ten designs from ICCAD2013 CAD Contest \cite{OPC-ICCAD2013-Banerjee}. 
Because in most cases model-based OPC runs faster than ILT, if we can efficiently predict the behavior of different OPC engines and hence choose the best one,
meanwhile the throughput of mask optimization flow can be significantly improved.
The observation, therefore, inspires the design of a heterogeneous OPC framework,
which adopts a deterministic machine learning model identifies the best OPC engine for a given design with negligible overhead.

As a case study, in this paper, we adopt two OPC engines that are based on ILT and compact model respectively.
We adopt the same training design set as used to train GAN-OPC \cite{OPC-DAC2018-Yang} which are fed into an ILT engine \cite{OPC-DAC2014-Gao} and a model-based OPC \cite{OPC-DATE2015-Kuang}.
Each design in the training set is labeled according to which OPC engine behaves best.
For the classification neural networks, we use the same architecture as in \cite{HSD-TCAD2019-Yang}.
Layout images are also converted to DCT format accordingly.

We evaluate the proposed framework using ten designs from ICCAD2013 CAD Contest \cite{OPC-ICCAD2013-Banerjee}.
Each design is fed into the trained CNN model before going through the mask optimization stage.
CNN predicts which OPC engine behaves better on the given design.
Detailed results are listed in \Cref{tab:rst}, where
``MB-OPC'', ``ILT'' and ``H-OPC'' list the results of model-based OPC, inverse lithography technique-based OPC and the proposed heterogeneous OPC respectively.
In the table, column ``ID'' represents 10 designs included in the benchmark suite, 
columns ``MSE'' indicate the mean square error between the simulated wafer image and the design for each OPC solution,
and columns ``Time'' list the mask optimization runtime of each design using three solutions.
As can be seen, the proposed heterogeneous OPC framework can assign better OPC engines to 8 out of ten designs in the benchmark suit, 
which hence results in better mask optimization performance with average MSE reduced by $\sim3\%$.
Also, the trade-off on runtime overhead is more balanced with the help of a deterministic learning model.

\section{Conclusion and Discussion}
\label{sec:conclu}

In this paper, we study recent advances of machine learning techniques on VLSI mask optimization problems.
We show that both deterministic and generative machine learning models assist to manufacturing-friendly layout design.
The former helps to identify process weak regions in a design and can speed-up OPC by circumventing costly lithography simulation.
The latter focuses on generation of directly printable masks.
Observing the importance of legacy OPC engines in machine learning-based solutions,
we propose a new methodology that a machine learning model facilitates modern OPC flow.
A deterministic classification model is designed to identify the best OPC engine for a given design with negligible computing overhead.
We hope the study can motivate deeper explorations of machine learning solutions for VLSI mask optimization,
which should not only include research on machine learning-based OPC engine itself but should also dig into a flow control level.

\section*{acknowledgment}
This work is supported by The Research Grants Council of Hong Kong SAR (Project No.~CUHK24209017).

{
    \footnotesize
	\bibliographystyle{IEEEtran}
	\bibliography{./ref/Top-sim,./ref/HSD,./ref/DFM,./ref/DL,./ref/MPL}
}

\end{document}